\documentclass[11pt]{article}
\usepackage{amsmath,amssymb,amsthm,hyperref,multirow}
\usepackage{enumitem}

\newcommand{\R}{\mathbb{R}}

\usepackage[normalem]{ulem}
\usepackage{textcomp}

\usepackage{booktabs}

\usepackage{color}
\usepackage{ulem}

\newsavebox{\measurebox}
\usepackage[numbers,sort&compress]{natbib}
\usepackage{authblk}
\usepackage{algorithm}
\usepackage{algpseudocode}
\usepackage[table]{xcolor}
\usepackage{graphicx}
\usepackage{epstopdf}
\usepackage{subfig}
\usepackage{dsfont, physics,stmaryrd,tabu}
\usepackage{enumitem}
\usepackage{mathtools}

\theoremstyle{plain}
\newtheorem{theorem}{Theorem}[section]
\newtheorem*{theorem*}{Theorem}

\theoremstyle{definition}

\theoremstyle{remark}

\newtheorem{remark}[theorem]{Remark}

\numberwithin{equation}{section}
\numberwithin{algorithm}{section}
\numberwithin{figure}{section}
\numberwithin{table}{section}
\numberwithin{theorem}{section}

\usepackage{array}
\newcolumntype{P}[1]{>{\centering\arraybackslash}p{#1}}
\title{N\MakeLowercase{eu}PDE: Neural Network Based Ordinary and Partial Differential Equations for Modeling Time-Dependent Data}
\author{Yifan Sun} \author{Linan Zhang} \author{Hayden Schaeffer}
\affil{Department of Mathematical Sciences, Carnegie Mellon University, Pittsburgh, PA 15213.}
\date{~}
\begin{document}

\maketitle
\begin{abstract}
We propose a neural network based approach for extracting models from dynamic data using ordinary and partial differential equations. In particular, given a time-series or spatio-temporal dataset,  we seek to identify an accurate governing system which respects the intrinsic differential structure. The unknown governing model is parameterized by using both (shallow) multilayer perceptrons and nonlinear differential terms, in order to incorporate relevant correlations between spatio-temporal samples. We demonstrate the approach on several examples where the data is sampled from various dynamical systems and give a comparison to recurrent networks and other data-discovery methods. In addition, we show that for MNIST and Fashion MNIST, our approach lowers the parameter cost as compared to other deep neural networks.\end{abstract}

\section{Introduction}\label{introduction}

Modeling and extracting governing equations from complex time-series can provide useful information for analyzing data.  An accurate governing system could be used for making data-driven predictions, extracting large-scale patterns, and uncovering hidden structures in the data. In this work, we present an approach for modeling time-dependent data using differential equations which are parameterized by shallow neural networks, but retain their intrinsic (continuous) differential structure.  

 For time-series data, recurrent neural networks (RNN) is often employed for encoding temporal data and forecasting future states. Part of the success of RNN are due to the internal memory architecture which allows these networks to better incorporate state information over the length of a given sequence. Although widely successful for language modeling, translation, and speech recognition, their use in high-fidelity scientific computing applications is limited. One can observe that a sequence generated by an RNN may not preserve temporal regularity of the underlying signals (see, for example \cite{chen2018neural} or Figure \ref{fig:lorenz_test_RNN}) and thus may not represent the true continuous dynamics.

For imaging tasks, deep neural networks (DNN) such as ResNet \cite{he2015resnet, he2016identity}, FractalNet \cite{larsson2016fractal}, and DenseNet \cite{huang2016dense} have been successful in extracting complex hierarchical spatial information.  These networks utilize intra-layer connectivity to preserve feature information over the network depth. For example, the ResNet architecture uses convolutional layers and skip connections. The hidden layers take the form $x^{n+1}=x^{n}+F(x^n,\theta)$ where $x^n$ represents the features at layer $n$ and $F$ is a convolutional neural network (or more generally, any universal approximator) with trainable parameters $\theta$. The evolution of the features over the network depth is equivalent to applying the forward Euler method to the ordinary differential equation (ODE): $\dot{x}=F(x,\theta)$. The connection between ResNet's architecture, numerical integrators for differential equations, and optimal control has been presented in \cite{weinan2017proposal, lu2017beyond, haber2017inverse, ruthotto2018pde, zhang2018forward}.

Recently, DNN-based approaches related to differential equations have been proposed for data mining, forecasting, and approximation. Examples of algorithms which use DNN  
for learning ODE and PDE include: learning from data using a PDE-based network \cite{long2017pde, long2018pde},
deep learning for advection equations \cite{de2017deep}, approximating dynamics using ResNet with recurrent layers \cite{qin2018data}, and learning and modeling solutions of PDE using networks  \cite{raissi2017physics}. Other approaches for learning governing systems and dynamics involve sparse regularizers ($\ell_0$ or hard-thresholding approaches in \cite{brunton2016discovering, rudy2017data, schaeffer2017sparse, schaeffer2017bifurcation} and $\ell_1$ problems in \cite{tran2017exact, schaeffer2017learning, schaeffer2018extracting})
or models based on Gaussian processes \cite{raissi2017machine, raissi2018hidden}.

Note that in \cite{long2017pde, long2018pde} it was shown that adding more blocks of the PDE-based network improved (experimentally) the model's predictive capabilities. Using ODEs to represent the network connectivity, \cite{chen2018neural} proposed a `continuous-depth' neural network called ODE-Net. Their approach essentially replaces the layers in ResNet-like architectures with a trainable ODE. In \cite{chen2018neural}, the authors state that their approach has several advantages, including the ability to better connect `layers' due to the continuity of the model and a lower memory cost when training the parameters using the adjoint method. The adjoint method proposed in  \cite{chen2018neural} may not be stable for a general problem. In \cite{gholami2019anode}, a memory efficient and stable approach for training a neural ODE was given.

\subsection{Contributions of this Work.} 
We present a machine learning approach for constructing approximations to governing equations of time-dependent systems that blends physics-informed candidate functions with neural networks. In particular, we construct a network approximation to an ODE which takes into account the connectivity between components (using a dictionary of monomials) and the differential structure of spatial terms (using finite difference kernels). If the user has prior knowledge on the structure or source of the data, i.e. fluids, mechanics, etc., one can incorporate commonly used physical models into the dictionary.  We show that our approach can be used to extract ODE or PDE models from time-dependent data, improve the spatial accuracy of reduced order models, and reduce the parameter cost for image classification (for the MNIST and Fashion MNIST datasets).

\section{Modeling Via Ordinary Differential Equations}\label{ode}

Given discrete-time measurements generated from an unknown dynamic process, we model the time-series using a (first-order) ordinary differential equation, $\dot{x}(t) = f(t, x(t))$, $x\in \R^d$ with $d\ge1$.
The problem is to construct an approximation to the unknown generating function $f$, i.e. we will learn networks $\text{net}(t,x)$ such that $\dot{x}\approx \text{net}(t,x)$. Essentially, we are learning a neural network approximation to the velocity field. Following the approach in \cite{chen2018neural}, the system is trained by a `continuous' model and the function $f$ is parameterized by multilayer perceptrons (MLP). Since a two-layer MLP may require a large width to approximate a generic (nonlinear) function $f$, we purpose a different parameterization.  Specifically, to better capture higher-order correlations between components of the data and to lower the number of parameters needed in the MLP (see for example, Figure~\ref{fig:ODE_noisy}), a dictionary of candidate inputs is added. Let $\mathcal{D}(t,x; p)$ be the collection (as a matrix) of the $p$th order monomial terms depending on $t$ and $x$, i.e. each element in $\mathcal{D}$ can be written as:

$$t^k x_1^{\ell_1}\cdots x_d^{\ell_d},  \ \ \text{for} \ \  0<k+\sum_i \ell_i \leq p.$$

One has freedom to determine the particular dictionary elements; however, the choice of monomial terms provides a model for the interactions between each of the components of the time-series and is used for model identification of dynamical systems in the general setting \cite{brunton2016discovering, tran2017exact, schaeffer2018extracting}. For simplicity, we will suppress the (user-defined) parameter $p$. 

In  \cite{brunton2016discovering, tran2017exact, schaeffer2018extracting, rudy2017data}, regularized optimization with polynomial dictionaries is used to approximate the generating function of some unknown dynamic process. When the dictionary is large enough so that the `true' function is in the span of the candidate space, the solutions produced by sparse optimization are guaranteed to be accurate. To avoid defining a sufficiently rich dictionary, we propose using an MLP (with a non-polynomial activation function) in combination with the monomial dictionary, so that general functions may be well-approximated by the network.  Note that the idea of using products of the inputs appears in other network architectures for example, the high-order neural networks  \cite{giles1987learning, shin1991pi}.
 
In many DNN architectures, batch normalization \cite{ioffe2015batch} or weight normalization \cite{salimans2016weight} are used to improve the performance and stability during training. For the training of NeuPDE, a simple (uniform) normalization layer, $N(x)$, is added between the input and dictionary layers, which maps $x$ to a vector in $[-1,1]^d$ (using the range over the all components). Specifically, let $M$ and $m$ be the maximum (and minimum) value of the data, over all components and samples and define the vector $N(x)$ as:
$$N(x) := 2 \frac{x-m\, \mathrm{1}_d}{M-m} -\mathrm{1}_d \in [-1,1]^d$$
This normalization is applied to each component uniformly and enforces that each component of the dictionary is bounded by one (in magnitude). We found that this normalization was sufficient for stabilizing training and speeding up optimization in the regression examples. Without this step, divergence in the training phase was observed.

To train the network: let $\theta$ be the vector of learnable parameters in the MLP layer, then the optimization problem is:
\begin{align}\label{eqn:train}
\min_{\theta} \quad &\sum_{i=1}^N \ L(x(t_i)) + \beta_1 r(\theta) + \frac{\beta_2}{2} \int_{t_0}^{t_N} \left\| \dot{x}(\tau)\right\|_{\ell^2}^2 d\tau\\
&s.t. \quad x(t_0)=x_0, \quad \dot{x} = F(\mathcal{D}(N(x)),\theta) \nonumber
\end{align}
where $\beta_1, \beta_2>0$ are regularization parameters set by the user and $F$ is an MLP.  Specifically, let $\sigma$ be a smooth activation function, for example, the exponential linear unit (ELU) 
\begin{align*}
\sigma_{\text{ELU}}(x)= 
\begin{cases}
&e^x-1, \ \ x\geq 0\\
&x, \hspace{1.05cm} x<0
\end{cases}
\end{align*}
or the hyperbolic tangent, tanh, which will be sufficiently smooth for integration using Runge-Kutta schemes. The right-hand side of the ODE is parameterized by a fully connected layer - activation layer - fully connected layer, i.e.
$$F(z,\theta) := A_2 \,  \sigma(\, A_1 z +b_1) + b_2$$
where $\theta=\text{vect}(A_1,A_2,b_1,b_2)$, i.e. the vectorization of all components of the matrices $A_1$ and  $A_2$ and biases $b_1$ and $b_2$. Therefore, the first layer of the MLP in the form $F(\mathcal{D}(N(x)),\theta)$ takes a linear combination of candidate functions (applied to normalized data). Note that the dictionary does not include the constant term since we include a bias in the first fully connected layer. The function $r$ is a regularizer on the parameters (for example, the $\ell^1$ norm) and the time-derivative is penalized by the $L^2$ norm. When used, the parameters are set to $\beta_1=10^{-4}$ and $\beta_2 =10^{-5}$ (no tuning is performed).  

The constraints in Eqn. \eqref{eqn:train} are written in continuous-time, i.e. the value of $x(t)$ is defined by the ODE and thus can be evaluated at any time $t\in [t_0, t_N]$.  For a given set of parameters $\theta$, the values $x(t_i)$ are obtained by numerical integration (for example, using a Runge-Kutta scheme). To optimize Eqn. \eqref{eqn:train} using a gradient-based method, the back-propagation algorithm or the adjoint method (following \cite{chen2018neural}) can be used. The adjoint method requires solving the ODE (and its adjoint) backward-in-time, which can lead to numerical instabilities. Following the approach in \cite{gholami2019anode}, checkpointing can be used to mitigate this issue.

For all experiments, we take the `\textit{discretize-then-optimize}' approach. The objective function, Eqn. \eqref{eqn:train}, is discretized as follows:

\begin{align}\label{eqn:trainD}
\min_{\theta} \quad &\sum_{i=1}^N \ L(x(t_i)) + \beta_1 r(\theta) + \frac{\tilde{\beta}_2}{2} \sum_{i=0}^{N-1} \left\| x(t_{i+1}) - x(t_i)\right\|_{\ell^2}^2\\
&s.t. \quad x(t_0)=x_0, \quad x(t_{i}) = \Phi^{(i)}(x(t_0), F(\mathcal{D}(N(-)),\theta))  \nonumber
\end{align}
where $\Phi^{(i)}$ is an ODE solver (i.e. a Runge-Kutta scheme) applied $i$-times,  $\tilde{\beta}_2$ is $\beta_2$ rescaled by the time-step, and the time-derivative is discretized on the time-interval with the integral approximated by piece-wise constant quadrature. The constraint that the ODE  $\dot{x} = F(\mathcal{D}(N(x)),\theta)$ is satisfied at every time-stamp has been transformed to the constraint that the sequence $x(t_i)$ for $0\leq i \leq N$ is generated by the forward evolution of an ODE solver. The ODE solver takes (as its inputs) the initial data $x(t_0)$ and the function $F$ that defines the RHS of the ODE. Note that the ODE solver can be `sub-grid' in the sense that, over a time interval $[t_i,t_{i+1}]$, we can set the solver to take multiple (smaller) time-steps. This will increase storage cost needed for back-propagation; however, taking multiple time-steps can better resolve complex dynamics embedded by $F \circ \mathcal{D}$ (see examples below). Additionally, the time-derivative regularizer helps to control the growth of the generative model, which yields control over the solution $x(t)$ and its regularity.

Eqn. \eqref{eqn:trainD} is solved using the Adam algorithm \cite{kingma2014adam} with temporal mini-batches. Each mini-batch is constructed by taking a random sampling of the time-stamps $t_i$ for $i=1, \cdots, N-k$ and then collecting all data-points from $x(t_i)$ to $x(t_{i+k})$ for some $k>0$.  This can be easily extended to multiple time-series, by sampling over each trajectory as well.  Note that, in experiments, taking non-overlapping sub-intervals did not change the results. The back-propagation algorithm \cite{maclaurin2015autograd} applied to each of the subintervals $[t_i, t_{i+k}]$ can be done in parallel.   For all of the regression examples, we set $L(x(t_i)):=||x(t_i)-\tilde{x}_i||^2_2$ where $\left\{x_i \right\}_{i=0}^N$ is the given (sequential) data. For the image classification examples, we used the standard cross-entropy for $L$.
\begin{remark}
\emph{Layers}: The right-hand side of the ODE is parameterized by one set of parameters $\theta$. Therefore, in terms of DNN layers, we are technically only training one ``layer''. However, changes in the structure between time-steps are taken into account by the time-dependence, i.e. the dictionary terms $\mathcal{D}$ that contain $t$. Thus, we are embedding multiple-layers into the time-dependent dictionary.
\end{remark}

\begin{remark}
`\emph{Continuous-depth}': Eqn. \eqref{eqn:trainD} is fully discrete when using a  Runge-Kutta solver for $\Phi$ and its gradient can be calculated using the back-propagation algorithm. If we used an adaptive ODE solver, such as Runge-Kutta 45, the forward propagation would generate a new set of time-stamps (which always contain the time-stamps $\{t_i\}_{i=0}^N$) in order to approximate the forward evolution $\dot{x} = F(\mathcal{D}(N(x)),\theta)$, given an error tolerance and a set of parameters $\theta$. We tested the continuous-depth versions, using back-propagation and the adjoint method (see Appendix A). Although the backward evolution of the ODE constraint may not be well-posed (i.e. numerical unstable or unbounded), our experiments lead to similar results. This could be due to the temporal mini-batches which enforce short-integration windows. Following \cite{gholami2019anode}, a checkpointing approach should be used, which would also control the memory footprint. It should be noted that, the adjoint approach was tested for the ODE examples but not for PDE examples, since the PDE examples are not time-reversible and will lead to backward-integration issues.

In addition, when the network is discrete, one may still consider it as a `continuous-depth' approximation, since the underlying approximation can be refined by adding more time-stamps, without the need to retrain the MLP.
\end{remark}

\subsection{Autonomous ODE.}
\begin{figure}[t!]   
    \centering
    \subfloat[Degree 2, 263 param.]{
    \includegraphics[scale=0.6]{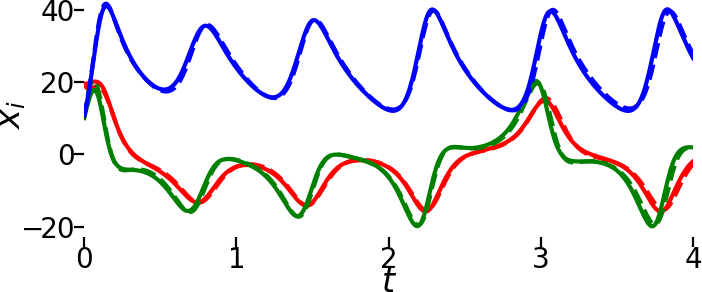}} 
    \subfloat[Degree 1, 269 param.]{
    \includegraphics[scale=0.6]{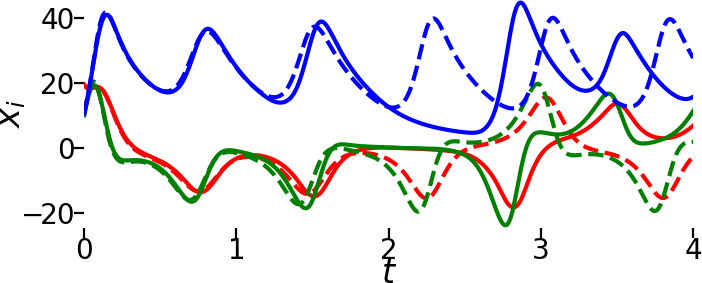}} 
    \subfloat[Degree 1, 703 param.]{
    \includegraphics[scale=0.6]{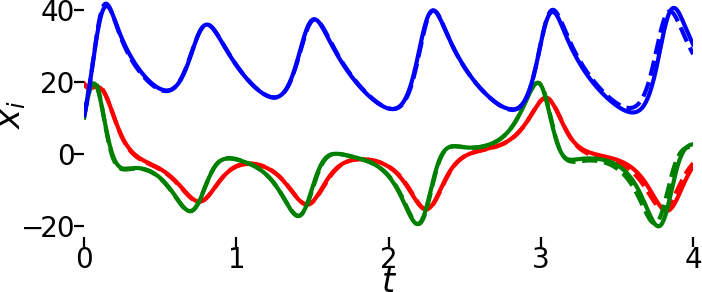}} 
    \caption{Time-series data generated by the 3d Lorenz system and the corresponding learned processes using our approach.  The original data (dashed) and the learned series (solid) are plotted, where red, green, and blue curves correspond to the  $x_1$, $x_2$, and $x_3$ components, respectively.}
       \label{fig:ODE_notests}
\end{figure}
When the time-series data is known to be autonomous, i.e. the ODE takes the form $\dot{x} = f(x)$, one can drop the $t$-dependency in the dictionary. In this case, the monomials take the form $x_1^{\ell_1}\cdots x_d^{\ell_d}$. We train this model by minimizing:
\begin{align}\label{eqn:trainDAuto}
\min_{\theta} \quad &\sum_{i=0}^{N-1} \ ||x(t_i)-\tilde{x}_i||_{\ell^2}^2 + \beta_1 \|\theta\|_{\ell^1} + \frac{\tilde{\beta}_2}{2} \sum_{i=1}^N \left\| x(t_{i+1}) - x(t_i)\right\|_{\ell^2}^2\\
&s.t. \quad x(t_0)=\tilde{x}_0, \quad x(t_{i}) = \Phi^{(i)}(x(t_0), F(\mathcal{D}(N(-),\theta)) \nonumber
\end{align}
where $\tilde{x}_i$ is the given data (corrupted by noise) over the time-stamps $t_i$ for $0\leq i \leq N$. The true governing equation is given by the 3d Lorenz system:
    \begin{align}
    \begin{cases}
        \dot{x}(t) &= 10 (y-x) \\
        \dot{y}(t) &= x(28-z)-y \\
        \dot{z}(t) &= xy - 8z/3
        \end{cases}\label{eq:lorenz}
    \end{align}  
which emits chaotic trajectories.

In Figure \ref{fig:ODE_notests}(a), we train the model with 20 hidden nodes per layer using a quadratic dictionary, i.e. there are 9 terms in the dictionary,  $A_1\in \mathbb{R}^{20 \times 9}$ with 20 bias parameters, $A_2\in \mathbb{R}^{3 \times 20}$ with 3 bias parameters,  for a total of 263 trainable parameters. The solid curves are the time-series generated by a forward pass of the trained model. The learned system generates a high-fidelity trajectory for the first part of the time-interval. In Figure \ref{fig:ODE_notests}(b-c), we investigate the effect of the degree in the dictionary. In Figure \ref{fig:ODE_notests}(b), using a degree 1 monomial dictionary with 38 hidden nodes per layers, i.e. 3 terms in the dictionary,  $A_1\in \mathbb{R}^{38 \times 3}$ with 38 bias parameters, $A_2\in \mathbb{R}^{3 \times 38}$ with 3 bias parameters (for a total of 269 trainable parameters), the generated curves trace a similar part of phase space, but are point-wise inaccurate.   By increasing the hidden nodes to 100 per layer (3 terms in the dictionary,  $A_1\in \mathbb{R}^{100 \times 3}$ with 100 bias parameters, $A_2\in \mathbb{R}^{3 \times 100}$ with 3 bias parameter, for a total of 703 trainable parameters), we see in Figure \ref{fig:ODE_notests}(c) that the method (using a degree 1 dictionary) is able to capture the correct point-wise information (on the same order of accuracy as Figure \ref{fig:ODE_notests}(a)) but requires more than double the number of parameters.

\subsection{Non-Autonomous ODE and Noise.}
\begin{figure}[b!]
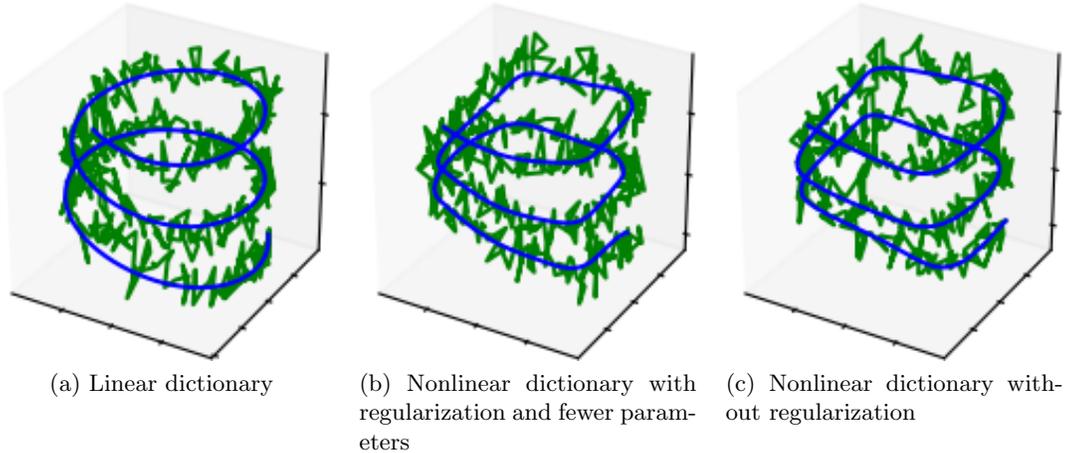

    \centering
    \subfloat[Linear dictionary]{
    \includegraphics[scale=1.1]{spiral3D_nodict.png}}\quad
    \subfloat[Nonlinear dictionary with regularization and fewer parameters]{
    \includegraphics[scale=1.1]{spiral3D_lessparam.png}}\quad
    \subfloat[Nonlinear dictionary without regularization]{
    \includegraphics[scale=1.1]{spiral3D_dict.png}} 
      \caption{Extracting and modeling of a nonlinear time-dependent spiral. Noisy data is given in green and learned model is given in blue. }
    \label{fig:ODE_noisy}
\end{figure}

To investigate the effects of noise and regularization, we fit the data to a non-linear spiral:
   \begin{align}
    \begin{cases}
        \dot{x}(t) &= 2 y(t)^3 \\
        \dot{y}(t) &= -2 x(t)^3\\
        \dot{z}(t) &= \frac{1}{4} + \frac{1}{2}\sin(\pi\,t)
        \end{cases}\label{eq:NLspiral}
    \end{align} 
    corrupted by noise.
  The third coordinate of Eqn.~\eqref{eq:NLspiral} is time-dependent, which can be challenging for many recovery algorithms. This is partly due to the redundancy introduced into the dictionary by the time-dependent terms. To generate Figure~\ref{fig:ODE_noisy}, we set: 
\begin{itemize}
\item[(a)] the dictionary degree to 1, with 4 terms, $A_1\in \mathbb{R}^{46 \times 4}$ with 46 bias parameters, $A_2\in \mathbb{R}^{3 \times 46}$ with 3 bias parameters (371 trainable parameters in total);
\item[(b)] the degree to 4, with 69 terms, $A_1\in \mathbb{R}^{4 \times 69}$ with 4 bias parameter, $A_2\in \mathbb{R}^{3 \times 4}$ with 3 bias parameters (295 trainable parameters in total), and the regularization parameter to $\tilde{\beta}_2 = 10^{-5}$;
\item[(c)] the degree to 4, with 69 terms, $A_1\in \mathbb{R}^{5 \times 69}$ with 5 bias parameter, $A_2\in \mathbb{R}^{3 \times 5}$ with 3 bias parameters (368 trainable parameters in total).
\end{itemize}
For cases (b-c), we set the degree of the dictionary to be larger than the known degree of the governing ODE in order to verify that we do not overfit using a higher-order dictionary and that we are not tailoring the dictionary to the problem. In Figure \ref{fig:ODE_noisy}(a), the dictionary of linear monomials with a moderately sized MLP seems to be insufficient for capturing the true nonlinear dynamics. This can be observed by the over-smoothing caused by the linear-like dynamics. In Figure \ref{fig:ODE_noisy}(c), a nonlinear dictionary can fit the data and extract the correct pattern (the `squared'-off corners).  Figure~\ref{fig:ODE_noisy}(b) shows that we are able to decrease the total number of parameters and fit the trajectory within the same tolerance as (c) by penalizing the derivative. Both (b) and (c) have achieved a mean-squared loss under $0.015$.

\subsection{Comparison for Extracting Governing Models.}

 \textbf{Comparison with SINDy.} We compare the results of Figure~\ref{fig:ODE_noisy} with an approximation using the SINDy algorithm from \cite{brunton2016discovering} (theoretical results of convergence and relationship to the $\ell^0$ problem appear in \cite{zhang2019convergence}). These approaches differ, since the SINDy algorithm seeks to recover a sparse approximation to the governing system given one tuning parameter and is restricted to the span of the dictionary elements. To make the dictionary sufficiently rich, the degree is set to 4 as was done for Figure~\ref{fig:ODE_noisy} (b-c). Since the sparsity of the first two components is equal to one, we search over all parameter-space (up to 6 decimals) that yields the smallest non-zero sparsity. The smallest non-zero sparsity for the first component is 12 and for the second component is 3 with:
    \begin{align}
    \begin{cases}
        \dot{x}(t) &=  -4278.0+ 9426.6z-2204.6t-7594.0z^2+3381.8tz -351.1t^2+...\\   &2650.6z^3-1659.3tz^2 +285.5t^2z -339.0 z^4+264.1tz^3 -58.4t^2z^2\\
        \dot{y}(t) &= -79.1527+68.0904z-14.4623z^2\\
        \dot{z}(t) &=  -53.0629+  220.7608z -168.9863t -266.8949z^2  +289.1066tz  -32.6971t^2...\\
        &+ 127.4432 z^3 -161.9778tz^2 +31.9400t^2z  -21.1582z^4+   29.5282tz^3   -7.6042t^2z^2
        \end{cases}
    \end{align} 
 which is independent of $x$ and $y$ and does not lead to an accurate approximation to the nonlinear spiral. This is likely due to the level of noise present in the data and the incorporation of the time-component. 
 
\textbf{Comparison with LASSO-based methods.}
We compare the results of Figure~\ref{fig:ODE_noisy} with LASSO-based approximations for learning governing equations \cite{schaeffer2017learning}. The LASSO parameter is chosen so that the sparsity of the solution matches the sparsity of the true dynamics (with respect to a dictionary of degree 4). In addition, the coefficients are `debiased' following the approach in \cite{schaeffer2017learning}. The learned system is:
      \begin{align}
    \begin{cases}
        \dot{x}(t) &= 1.8398 y^3   \\
        \dot{y}(t) &= -1.9071x^3\\
        \dot{z}(t) &=    -0.1749x^2y-0.0058 t^2x-0.0008t^2x^2
        \end{cases}
    \end{align} 
 which matches the profile of the data in the $(x,y)$-plane; however, it does not predict the correct dynamics for $z$ (emitting seemingly periodic orbits). While the LASSO-based approach better resolves the state-space dependence, it does not correctly identify the time-component.

\begin{figure}[h!]
    \centering{
    \includegraphics[scale=0.8]{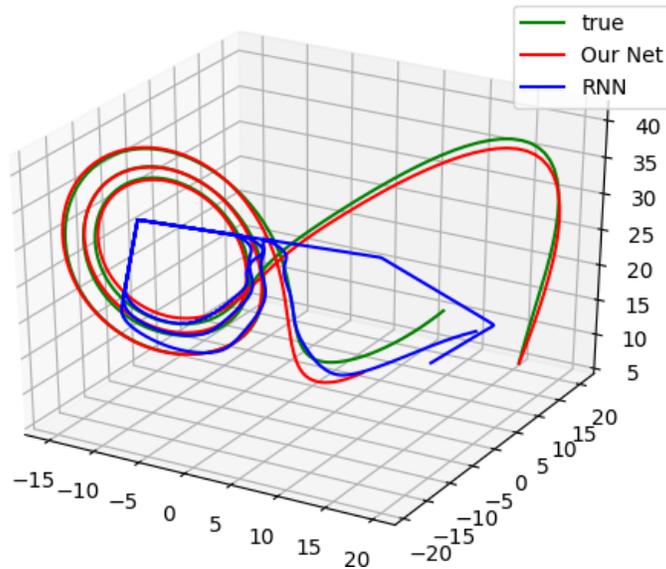}} \caption{Comparing dynamics between the true (green) time-series,  the solution generated by our network (red), and a solution generated by an RNN (blue) with the same number of parameters.}
    \label{fig:lorenz_test_RNN}
\end{figure}

 \textbf{Comparison with RNN.} In Figure~\ref{fig:lorenz_test_RNN} the Lorenz system (see Figure~\ref{fig:ODE_notests}) is approximated by our proposed approach and a standard LSTM (RNN), with the same number of parameters. Although the RNN learns internal hidden states, the RNN does not learn the correct regularity of the trajectories thus leading to sharp corners. It is worth noting that, in experiments, as the number of parameters increases, both the RNN and our network will produce sequences that approach the true time-series.

\subsection{ODE from Low-Rank Approximations }
\label{pod}

For certain spatio-temporal systems, reduced-order models can be used to transform complex dynamics into low-dimensional time-series (with stationary spatial modes). One of the popular methods for extracting the spatial modes and identifying the corresponding temporal-dynamics is the dynamic mode decomposition (DMD) introduced in \cite{schmid2008dynamic}. The projected DMD method \cite{schmid2010dynamic} makes use of the SVD approximation to construct the modes and the linear dynamical system. Another reduced-order model, known as the proper orthogonal decomposition (POD) \cite{holmes2012turbulence}, can be used to construct spatial modes which best represent a given spatio-temporal dataset. The projected DMD and the POD methods leverage low-rank approximations to reduce the dimension of the system and to construct a linear approximation to the dynamics (related to the spectral analysis of the Koopman operator), see  \cite{DynamicModeDecomposition} and the citations within.

We apply our approach to construct a neural network approximation to the time-series generated by a low-rank approximation of the von K\'arm\'an vortex sheet. We explain the construction for this example here but for more details, see \cite{DynamicModeDecomposition}. Given a collection of measurements $\{u(x,y, t_i)\}_{i=0}^{N-1}$, where $(x,y)\in \Omega \subset \mathbb{R}^2$ are the spatial coordinates and $t_i$ are the time-stamps,  define $X$ as the matrix whose columns are the vectorization of each $u(x,y,t_i)$, i.e. $X_{-,i}: = \text{vect}(u(x,y,t_i))$ and $X \in \mathbb{R}^{m \times N}$ where $m$ is the number of grid points used to discretize $\Omega$.  The SVD of the data is given by $X=U\Sigma V^*$, where $U \in\mathbb{R}^{ m \times m}$ and $V\in\mathbb{R}^{ N \times N}$ are unitary matrices and $\Sigma \in\mathbb{R}^{ m \times N}$ is a diagonal matrix.
The best $r$-rank approximation of $X$ is given by $X_r := U_r \Sigma_r V_r^*$ where $\Sigma_r\in\mathbb{R}^{ r \times r}$ is the restriction of $\Sigma$ to the top $r$ singular values and $U_r\in\mathbb{R}^{ m \times r}$ and $V_r\in\mathbb{R}^{ N \times r}$ are the corresponding singular vectors. The columns of the matrix $U_r$ represent the $r$ spatial modes that can be used as a low-dimensional representation of the data. In particular, we define the vector $\alpha \in \mathbb{R}^r$ by the projection of the data (i.e. the columns of $X$) onto the span of $U_r$, that is:
$$\tilde{\alpha}(t_i):=U_r^*X_{-,i+1}.$$
Thus, we can construct the time-stamps $\tilde{\alpha}(t_i)$ from the measurements $X$ and can train the system using a version Eqn. \eqref{eqn:train} with the constraint that the ODE is of the form: 
$$\dot{\alpha}=A_0\alpha+f(\alpha).$$ 
The additional matrix $A_0\in \mathbb{R}^{r \times r}$ resembles the standard linear structure from the DMD approximation and the function $f$ can be seen as a nonlinear closure for the linear dynamics. The function $f$ is approximated, as before, by $F(\mathcal{D}(N(-), \theta)$.   To train the model, we minimize:
\begin{align}
\min_{\theta} \quad &\sum_{i=1}^{N-1} \ ||\alpha(t_i)-U_r^*X_{-,i+1}\|_{\ell^2}^2 + \beta_1 \|\theta\|_{\ell^1} + \frac{\tilde{\beta}_2}{2} \sum_{i=0}^{N-2} \left\| \alpha(t_{i+1}) - \alpha(t_i)\right\|_{\ell^2}^2\\
&s.t. \quad \alpha(t_0)=U_r^*X_{-,1}, \quad \alpha(t_{i}) = \Phi^{(i)}(\alpha(t_0), G( \mathcal{D}(N(-)),\theta)) \nonumber
\end{align}
where $G(\mathcal{D}(N(\alpha), \theta)= A_0\alpha+F( \mathcal{D}(N(\alpha),\theta)$ and $\theta$ also       includes the trainable parameters from $A_0$. Note that, to recover an approximation to the original measurements $u(x,y, t_i)$, the vector $U_r \alpha(t_i)$ is mapped back to the correct spatial ordering (inverting the vectorization process).

\begin{figure}[h!]
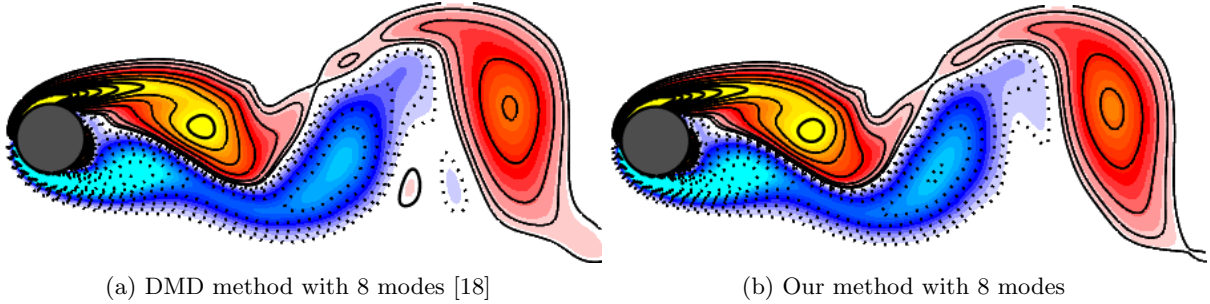

    \centering
    \subfloat[DMD method with 8 modes \cite{kutz2016dynamic}]{
    \includegraphics[scale=0.5]{DMD8.png}}
    \subfloat[Our method with 8 modes]{
    \includegraphics[scale=0.5]{OurROM8.png}} 
      \caption{Learning reduced-order dynamics from fluid simulations.}
    \label{fig:DMD}
\end{figure}

In Figure~\ref{fig:DMD}, our approach with an 8 mode decomposition is compared to an 8 mode DMD approximation. The DMD approximation in Figure~\ref{fig:DMD}(a) introduces two erroneous vortices near the bottom boundary. Our approach matches the test data with higher accuracy, specifically, the relative $L^2$ error between our generated solution at the terminal time is $0.049$ compared to DMD's relative error of $0.060$. It is worth noting that this example shows the benefit of the additional term $f(\alpha)$ in the low-mode limit; however, using more modes, the DMD method becomes a very accurate approximation. Unlike the standard DMD method, our model does not require the data to be uniformly spaced in time.

\section{Partial Differential Equations}
\label{pde}
 A general form for a first-order in time, $a$-th order in space, nonlinear PDE is: 
 $$u_t=G(t,x,u,D u, D^2 u,\cdots, D^a u),$$
where $D^i u$ denotes the collection of all $i$-th order spatial partial derivatives of $u$ for $1\leq i \leq a$. We form a dictionary $\mathcal{D}([t, x,u,Du, D^2u,\cdots, D^{a} u])$ as done in Sec. \ref{ode}, where the monomial terms now apply to $t$, $x$, $u$, and $D^i u$ for $1\leq i\leq a$. The spatial derivatives $D^i u$ as well as $u_t$ can be calculated numerically from data using finite differences. We then use an MLP, $F$, to parametrize the governing equation:
\begin{align}\label{eqn:PDEform}
    u_t = F\bigg(\mathcal{D}([t,x,u,Du, D^2u,\cdots, D^{\alpha} u]), \theta\bigg),
\end{align}
see also \cite{schaeffer2017learning, rudy2017data}. In particular, the function $F$ can be written as:
\begin{align}\label{eqn:PDENet}
    F(z,\theta)=K_2(\sigma(K_1(z)+b_1))+b_2
\end{align}
where $K_1$ and $K_2$ are collections of $1\times 1$ convolutions, $b_1$ and $b_2$ are biases, $\theta$ are all the parameters from $K_\ell$ and $b_\ell$, and $\sigma$ is ELU activation function. The input channels are the monomials determined by $t$, $x$, $u$, and $D^iu$, where $t$ is extended to a constant 2d array. The first linear layer maps the dictionary terms to multiple hidden channels, each defined by their own $1\times 1$ convolution. Thus, each hidden channel is a linear combination of input layers.  Then we apply the ELU activation, followed by a $1 \times 1$ convolution, which is  equivalent to taking linear combinations of the activated hidden channels. Note that this differs from \cite{schaeffer2017learning, rudy2017data} in several ways. In the first linear layer, our network uses multiple linear combinations rather than the single combination as in \cite{schaeffer2017learning, rudy2017data}. Additionally, by using a (nonlinear) MLP we can approximate a general function on the coordinates and derivative; however, previous work defined approximations that model functions within the span of the dictionary elements.

To illustrate this approach, we apply the method to two examples: a regression problem using data from a 2d Burgers' simulation (with noise) and the image classification problem using the MNIST and MNIST-Fashion datasets.

\subsection{Burgers' Equation}

We consider the 2d Burgers' equation, 
$$u_t+0.5\, \text{div}\left(u^2\right)=0.01 \Delta u.$$
The training and test data are generated on $(t,x,y)\in [0,0.015]\times [0,1]^2$, with time-step $\Delta t=1.5\times 10^{-5}$ and a $32\times 32$ uniform grid. To make the problem challenging, the training data is generated using a sine function in $x$ as the initial condition, while the test data uses a sine function in $y$ as the initial condition. We generate 5 training trajectories by adding noise to the initial condition.  Our training set is of size $[5,100,32,32]$ and our test data is of size $[1,100,32,32]$. To train the parameters we minimize:
\begin{align}
\min_{\theta} \quad &\sum_{i=0}^{N-1} \| u(x,y, t_i)-\tilde{u}(x,y,t_i)\|_{\ell^2(\Omega_d)}^2 + \beta_1 \|\theta\|_{\ell^1} + \frac{\tilde{\beta}_2}{2} \sum_{i=1}^N   \| u(x,y, t_{i+1})-u(x,y, t_{i})\|_{\ell^2(\Omega_d)}^2\\
&s.t. \quad u(x,y, t_0)=\tilde{u}(x,y,t_0), \quad u(x,y, t_i)= \Phi^{(i)}(u(x,y, t_0), F(\mathcal{D}(N(-)),\theta) \nonumber
\end{align}
where $\Omega_d$ is a discretization of $\Omega$.

\textbf{Training, Mini-batching, and Results.}
The mini-batches used during training are constructed with mini-batches in time and the full-batch in space. For our experiment, we set a (temporal) batch size of 16 with a length of $3$, i.e. each batch is of size $[16,3,32,32]$ containing 16 short trajectories. The points are chosen at random, without overlapping. The initial points of each mini-batch are treated as the initial conditions for the batch, and our predictions are performed over the length of the trajectory. This is done at each iteration of the Adam optimizer with a learning rate of $0.1$.

\begin{figure}
   \centering
    \subfloat[Training data.]{
    \includegraphics[scale=2.2]{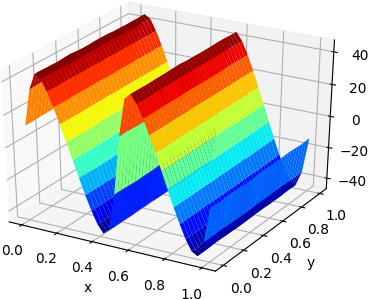}}
    \subfloat[Learned surface, test data.]{
    \includegraphics[scale=1.1]{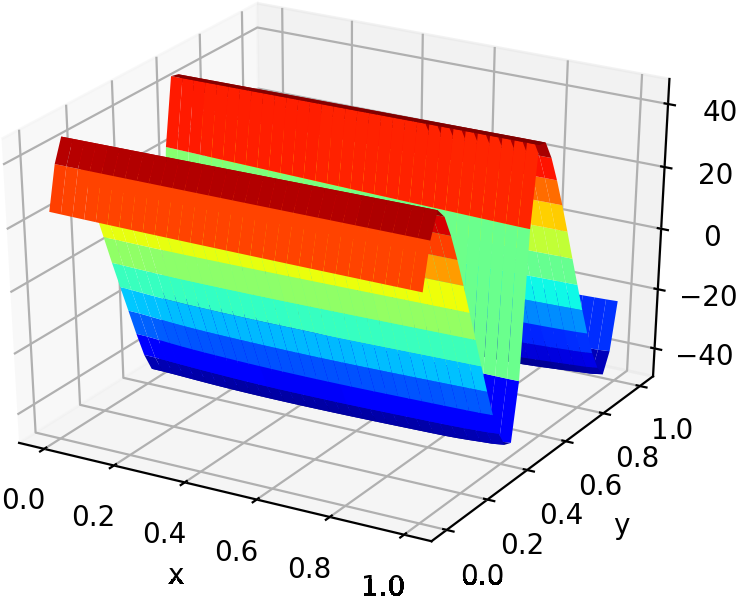}}\\
\caption{Burgers' Equation Example. The surfaces at the terminal time simulated by the NeuPDE, with (a) training data and (b) the learned surface on the test data.}
    \label{fig:burgers}
\end{figure}

In Figure~\ref{fig:burgers}, we take 2000 iterations of training, and evaluate our results on both the training and test sets. Each of the $1\times 1$ convolutional layers have 50 hidden units, for a total of 2301 learnable parameters. For visualization,  we plot the learned solution at the terminal time on both the training and test set. The mean-squared error  on the full training set is $0.005$ and on the test set is $3.6$ (for reference, the mean-squared value of the test data is over $1000$).

\subsection{Image Classification: MNIST Data.} 

Another application of our approach is in reducing the number of parameters in convolutional networks for image classification. We consider a linear (spatially-invariant) dictionary for Eqn.~\eqref{eqn:PDEform}. In particular, the right-hand side of the PDE is in the form of normalization, ReLU activation, two convolutions, and then a final normalization step. Each convolutional layer uses a $3 \times 3$ kernel of the form $\sum_{i=1}^6 a_i k_i,$ with 6 trainable parameters, where $k_i$ are $3 \times 3$ kernels that represent the identity and the five finite difference approximations to the partial derivatives $D_x$, $D_y$, $D_{xx}$, $D_{xy}$, and $D_{yy}$. In CNN \cite{lecun1998gradient, he2015resnet, he2016identity}, the early features (relative to the network depth) typically appear to be images that have been filtered by edge detectors \cite{ruthotto2018pde, zhang2018forward}. The early/mid-level trained kernels often represent edge and shape filters, which are connected to second-order spatial derivatives. This motivates us to replace the $3\times 3$ convolutions in ODE-Net \cite{chen2018neural} by finite difference kernels.

Result shows that even though the trainable set of parameters are decreased by a third (each kernel has 6 trainable parameters, rather than 9), the overall accuracy is preserved (see Table \ref{tableMNIST}). We follow the same experimental setup as in \cite{chen2018neural}, except that the convolutional layers are replaced by finite differences. 
We first downsample the input data using a downsampling block with 3 convolutional layers. Specifically, we take a $3\times 3$ convolutional layer with 64 output channels and then apply two $3\times 3$ convolutional layers with 64 output channels and a stride of 2. Between each convolutional layer, we apply batch-normalization and a ReLU activation function. The output of the downsampling block is of size $8\times 8$ with 64 channels. We then construct our PDE block using 6 `PDE' layers, taking the form:
\begin{align}\label{eqn:mnist ODE}
\dot{u}(t) = G(t,u,\theta_{i})\quad i\leq t\leq i+1,\ i\in\{0, \cdots,5\}.
\end{align}
We call each subinterval (indexed by $i$) a  PDE layer since it is the evolution of a semi-discete approximation of a coupled system of PDE (the particular form of the convolutions act as approximations to differential operators). The function $G$ takes the form:
\begin{align}
    G(u,\theta)=BN(K_2([t, BN(K_1([t,\sigma(N(u))]))]))
\end{align}
where $BN(x)$ is batch-normalization, $K_\ell$ is a collection of $3 \times 3$ kernels of the form $\sum_{i=1}^6 a_i k_i$, $\theta$ contains all the learnable parameters, and $\sigma(x)$ is the ReLU activation function. The PDE block is followed by batch-normalization, the ReLU activation function, and a 2d pooling layer. Lastly, a $64\times 10$ fully connected layer is used to transform the terminal state (activated and averaged) of the PDE blocks to a 10 component vector. 

For the optimization, the cross-entropy loss is used to compare the predicted outputs and the true label. We use the SGD optimizer with momentum set to 0.9. There are 160 total training epochs; we set the learning rate to 0.1 and decrease it by 1/10 after epoch 60, 100 and 140. The training is stopped after 160 epochs. All of the convolutions performed after the downsampling block are linear combinations of the 6 finite difference operators rather than the traditional $3\times 3$ convolution.

\begin{table}[h!] 
  \caption{Comparison Between Networks on MNIST}
  \label{tableMNIST}
  \centering
  \begin{tabular}{lll}
    \toprule
    \multicolumn{2}{c}{Method}                   \\
    \cmidrule(r){1-2}
    Name   &  \#Params. (M)     & Error($\%$) \\
    \midrule
    MLP \cite{lecun1998gradient} & 0.24  & 1.6     \\
    ResNet & 0.60  & 0.41     \\
    ODENet & 0.22  & 0.51     \\
    Our & 0.18  & 0.51     \\
    \bottomrule
  \end{tabular}
\end{table}

\subsection{Image Classification: Fashion MNIST.}
For a second test, we apply our network to the Fashion MNIST dataset. The Fashion MNIST dataset has 50000 training figures and 10000 test figures that can be classified into 10 different classes. Each data is of size $32\times 32$. For our experiment, we did not use any data augmentation, except for normalization before training. The structure of NeuPDE in this example differs from the MNIST experiment as follows: (1) we downsample the data once instead of twice and (2) after the 1st PDE block, which has 64 hidden units, we use a $3\times 3$ kernel with 64 input channels and 128 output channel, and then use one more PDE block with 128 hidden units followed by a $128\times 10$ fully connected layer. There are numerous reported test results on the Fashion MNIST dataset \cite{fashion2016}; we compare our result to ResNet18 \cite{fashion2016resnet} and a simple MLP \cite{MLP256}. The test results are presented in Table~\ref{tableFashion} and all algorithms are tested without the use of data augmentation. In both the MNIST and Fashion MNIST experiments, we show that the NeuPDE approach allows for less parameters by enforcing (continuous) structure through the network itself. 
\begin{table}[h]
  \caption{Comparison on Fashion MNIST}
  \label{tableFashion}
  \centering
  \begin{tabular}{lll}
    \toprule
    \multicolumn{2}{c}{Method}                   \\
    \cmidrule(r){1-2}
    Name   & \#Params. (M)     & Error ($\%$) \\
    \midrule
    MLP 256-128-100 & 0.248  & 11.67 \cite{MLP256}   \\
    ResNet18 & 2.78 & 8.0  \cite{fashion2016resnet}   \\
    Our & 0.38  & 7.6     \\
    \bottomrule
  \end{tabular}
\end{table}

\section{Discussion}

We propose a method for learning approximations to nonlinear dynamical systems (ODE and PDE) using DNN. The network we use has an architecture similar to ResNet and ODENet, in the sense that it approximates the forward integration of a first-order in time differential equation. However, we replace the forcing function (i.e. the layers) by an MLP with higher-order correlations between the spatio-temporal coordinates, the states, and derivatives of the states.  In terms of convolution neural networks, this is equivalent to enforcing that the kernels approximate differential operators (up to some degree). This was shown to produce more accurate approximations to complex time-series and spatio-temporal dynamics. As an additional application, we showed that when applied to image classification problems, our approach reduced the number of parameters needed while maintaining the accuracy. In scientific applications, there is more emphasis on accuracy and models that can incorporate physical structures. We plan to continue to investigate this approach for physical systems.

In imaging, one should consider the computational cost for training the networks versus the number of parameters used. While we argue that our architecture and structural conditions could lead to models with fewer parameter, it could be potentially slower in terms of training (due to the trainable nonlinear layer defined by the dictionary). Additionally, we leave the scalability of our approach for larger imaging data set, such as ImageNet, to future work. For larger classification problems, we suspect that higher-order derivatives (beyond second-order) may be needed.
Also, while higher-order integration methods (Runge-Kutta 4 or 45) may be better at capturing features in the ODE/PDE examples, more testing is required to investigate their benefits for imaging classification.

\section{Acknowledgements}
The authors would like to acknowledge the support of AFOSR, FA9550-17-1-0125 and the support of NSF CAREER grant $\#1752116$. We would like to thank Scott McCalla for providing feedback on this manuscript.

\bibliographystyle{plain}
\bibliography{PDElearn}
\clearpage
\newpage

\appendix
\section{Derivation of Adjoint Equations}

Let $\theta$ be the vector of learnable parameters (that parameterizes the unknown function $g$, which embeds all network features), then the training problem is:
\begin{align*}
\min_{\theta} \quad &\sum_{i=1}^N \ L(x(t_i)) + \beta_1 r(\theta) + \frac{\beta_2}{2} \int_{t_0}^{t_N} \left| \dot{x}\right|^2 d\tau\\
&s.t. \quad x(t_0)=x_0, \quad \dot{x} = g(x,t; \theta)
\end{align*}
where $\beta_1, \beta_2>0$ are regularization parameters set by the user. All subscripts with respect to a variable denote a partial derivative. We use the dot-notation for time-derivative for simplicity of exposition. The function $r$ is a regularizer on the parameters (for example, the $\ell_p$ norm) and the time-derivative is penalized by the $L^2$ norm.  Define the Lagrangian $\mathcal{L}$ by:
\begin{align*}
\mathcal{L} := \sum_{i=1}^N \ L(x(t_i)) + \beta_1 r(\theta) + \frac{\beta_2}{2} \int_{t_0}^{t_N} \left| \dot{x}\right|^2 d\tau-\int_{t_0}^{t_N} \lambda^T(\dot{x}- g(x,\tau; \theta)) d\tau
\end{align*}
where $\lambda(t)\in BV[t_0,t_N]$ is a time-dependent Lagrange multiplier. To apply gradient-based algorithms, the total derivative of the Lagrangian with respect to the trainable parameter must be calculated. Using integration by parts after differentiating with respect to $\theta$ yields:
\begin{align*}
\frac{d\mathcal{L}}{d\theta}&= \sum_{i=1}^N \ L_{x(t_i)}(x(t_i)) \ x_\theta(t_i)+ \beta_1 r_\theta(\theta) + \beta_2 \int_{t_0}^{t_N} \dot{x}^T \dot{x}_{\theta} \, d\tau- \sum_{i=1}^N\int_{t_{i-1}}^{t_i}\lambda^T(\dot{x}_{\theta}- g_x(x,\tau; \theta)x_{\theta}-g_\theta(x,\tau; \theta)) \ d\tau\\
&= \sum_{i=1}^N \ L_{x(t_i)}(x(t_i)) \ x_\theta(t_i)+ \beta_1 r_\theta(\theta) - \beta_2 \int_{t_0}^{t_N} \ddot{x}^T x_{\theta} \, d\tau +\beta_2 \dot{x}^T x_{\theta} \Big|_{t_0}^{t_N}  \\
&\quad + \sum_{i=1}^N\int_{t_{i-1}}^{t_i} \dot{\lambda}^T{x}_{\theta}+{\lambda}^Tg_x(x,\tau; \theta)x_{\theta}+{\lambda}^Tg_\theta(x,\tau;\theta) \ d\tau-  \sum_{i=1}^N \left(\lambda^T{x}_{\theta}\Big|_{t_{i-1}}^{t_i}\right)
\end{align*}
The initial condition $x(t_0)$ is independent of $\theta$, so $x_\theta(t_0)=0$. Define the evolution for $\lambda$ between any two time-stamps $[t_{i-1},t_i]$ by:
\begin{align*}
\dot{\lambda}^T(t) = -\lambda^Tg_x(x,t; \theta)+\beta_2 \ddot{x}^T
\end{align*}
then
\begin{align*}
&\frac{d\mathcal{L}}{d\theta}= \sum_{i=1}^N \ L_{x(t_i)}(x(t_i)) \ x_\theta(t_i)+ \beta_1 r_\theta(\theta) + \beta_2\dot{x}^T(t_N) x_{\theta}(t_N)  + \int_{t_0}^{t_N} {\lambda}^T g_\theta(x,\tau; \theta) \ d\tau-  \sum_{i=1}^N \left(\lambda^T{x}_{\theta}\Big|_{t_{i-1}}^{t_i}\right)
\end{align*}
To determine  $\lambda$ at $t_N$, we set $\lambda^T(t_N)=L_{x(t_N)}(x(t_N))+\beta_2\dot{x}^T(t_N)$ and at the right-endpoints of $[t_{i-1},t_{i}]$, we set:
$ \lambda(t^{+}_i)^T =  \lambda(t^{-}_i)^T+L_{x(t_i)}(x(t_i))$. The derivative of the Lagrangian with respect to $\theta$ becomes:
\begin{align*}
&\frac{d\mathcal{L}}{d\theta}=  \beta_1 r_\theta(\theta)   + \int_{t_0}^{t_N} {\lambda}^Tg_\theta(x,\tau; \theta) \ d\tau.
\end{align*}
Altogether, the evolution of $\lambda$ is define by:
\begin{align*}
\begin{cases}
\dot{\lambda}^T(t) = -\lambda^T g_x(x,t;\theta)+\beta_2 \ddot{x}^T,\quad \text{in} \quad [t_{i-1},t_i]\\
\lambda^T(t_N)=L_{x(t_N)}(x(t_N))+\beta_2\dot{x}^T(t_N)\\
 \lambda^T(t^{+}_i) =  \lambda^T(t^{-}_i)+L_{x(t_i)}(x(t_i)), \quad \text{for} \quad i=1, \cdots, N-1
 \end{cases}
\end{align*}
which can be re-written as:
\begin{align*}
\begin{cases}
\dot{\lambda}^T(t) = -\lambda^Tf_x(x,t)+\beta_2 \left( g_x(x,t; \theta) g(x,t;\theta) + g_t(x,t;\theta)\right)^T,\quad \text{in} \quad [t_{i-1},t_i]\\
\lambda^T(t_N)=L_{x(t_N)}(x(t_N))+\beta_2g(x(t_N),t_N;\theta)^T\\
 \lambda^T(t^{+}_i) =  \lambda^T(t^{-}_i)+L_{x(t_i)}(x(t_i)), \quad \text{for} \quad i=1, \cdots, N-1
 \end{cases}
\end{align*}
We augment the evolution for $\lambda(t)$ with $x(t)$, starting at $t=t_N$ and integrating backwards. The code follows the structure found in \cite{chen2018neural}.

\end{document}